\documentclass[runningheads]{llncs}
\usepackage{graphicx}
\usepackage{booktabs}
\usepackage[colorinlistoftodos]{todonotes}
\usepackage{pifont}

\usepackage[symbol]{footmisc}

\usepackage{enumitem,amssymb}
\newlist{todolist}{itemize}{2}
\setlist[todolist]{label=$\square$}

\begin{document}
\title{Light-weight spatio-temporal graphs for segmentation and ejection fraction prediction in cardiac ultrasound}

\titlerunning{GCNs for segmentation and EF prediction in cardiac ultrasound}
%
\author{Sarina Thomas \inst{1,\star} \and
Andrew Gilbert \inst{2,\star} \and
Guy Ben-Yosef \inst{3,\dagger}
}

\footnotetext[1]{S. Thomas and A. Gilbert contributed equally}
\footnotetext[2]{Corresponding author}

\authorrunning{S. Thomas, A. Gilbert et al.}
%
\institute{
University of Oslo, Oslo, Norway \and
GE Vingmed Ultrasound, Oslo, Norway \and
GE Research, Niskayuna, New York, USA \\ \email{guy.ben-yosef@ge.com}}

\maketitle              
\begin{abstract}
Accurate and consistent predictions of echocardiography parameters are important for cardiovascular diagnosis and treatment. In particular, segmentations of the left ventricle can be used to derive ventricular volume, ejection fraction (EF) and other relevant measurements. In this paper we propose a new automated method called EchoGraphs for predicting ejection fraction and segmenting the left ventricle by detecting anatomical keypoints. Models for direct coordinate regression based on Graph Convolutional Networks (GCNs) are used to detect the keypoints. GCNs can learn to represent the cardiac shape based on local appearance of each keypoint, as well as global spatial and temporal structures of all keypoints combined. We evaluate our EchoGraphs model on the EchoNet benchmark dataset. Compared to semantic segmentation, GCNs show accurate segmentation and improvements in robustness and inference run-time. EF is computed simultaneously to segmentations and our method also obtains state-of-the-art ejection fraction estimation. Source code is available online: \url{https://github.com/guybenyosef/EchoGraphs}.

\end{abstract}

\section{Introduction}
\label{sec:Introduction}
Heart failure continues to be the leading cause of hospitalization and death worldwide, lending an urgency to personalized and prospective care solutions \cite{Savarese2020}. Cardiovascular ultrasound (or echocardiography) is the most frequently used method for diagnosing heart disease due to its wide availability, low cost, real-time feedback, and lack of ionizing radiation.

Reduced left ventricle outflow -- an important marker of heart failure -- is measured by the ejection fraction (EF). EF describes the volumetric blood fraction pumped by the heart in each heart cycle and can be estimated by the ratio between the maximal volume (occurring at the end diastole phase of the cardiac cycle) and the minimal volume (occurring at the end systole phase). These volume estimates are typically derived from a segmentation of the myocardial border of the left ventricle. Segmentations of the myocardial border are also useful for many other downstream tasks such as foreshortening detection during acquisition \cite{Smistad2020} or as an initialization for strain measurement \cite{ostvik2018automatic}. Due to their frequent use and high prognostic value, accurate and robust automation of left ventricle segmentation and EF measurement are high priority tasks for modern echocardiography systems.

\subsection{Prior work} Several previous works have shown highly accurate automation results using deep learning for both left ventricle segmentation~\cite{Ouyang,Leclerc2019,Smistad2020,jafari2018,oktay2017anatomically,gilbert2021} and direct EF estimation~\cite{Ouyang,kazemi2020deep,reynaud2021ultrasound}. For segmentation, previous works have mostly relied on semantic segmentation, which outputs a pixel-wise classification of an input image. In some cases, extra modules were included to implicitly include shape constraints~\cite{oktay2017anatomically,Payer2019}, but in general these methods do not explicitly optimize the shape of interest. This is sub-optimal for a problem such as left ventricle segmentation because the shape is consistent between patients, and shape variations that do occur are critical for many diagnoses \cite{marciniak2021septal,baltabaeva2008regional}. Additionally, annotations are typically provided as keypoints, making keypoint regression a more natural form of learning.

The state-of-the-art for semantic segmentation in medical imaging is the nnU-Net \cite{isensee2021nnu}, a learning pipeline with a U-Net backbone. The success of the nnU-Net stems from its augmentations and automatic optimization of many network hyper-parameters based on dataset characteristics. 

Graph convolutional networks (GCNs) have been gaining popularity as a learning method to integrate multi-modal data that may not be grid-structured \cite{ahmedt2021graph}. However, GCNs are also ideally suited for predicting spatio-temporal key-point locations across time based on image or video features. For these problems the graph nodes represent the keypoints and the graph edges represent the learned relationship between points. GCNs have achieved state-of-the-art results in problems such as pose prediction in video \cite{wang2020motion} and also segmentation in several medical imaging problems  \cite{tian2020graph,gopinath2020graph}. In this work, we adapt graph convolutional approaches for echocardiography segmentation.

\subsection{Contributions:}
Given an echocardiography video loop of the left ventricle we use GCNs to demonstrate accurate automation of two relevant tasks for heart failure diagnosis: \textbf{(1) left ventricle segmentation} at end diastole (ED) and end systole (ES) phases of the cardiac cycle and \textbf{(2) ejection fraction estimation}. To our knowledge, this is the first work to apply graph convolutional approaches to these tasks and we show several advantages compared to prior works:
\begin{itemize}
    \item \textbf{Accuracy:} We demonstrate highly accurate segmentation and state-of-the-art ejection fraction prediction.
    \item \textbf{Robustness}: We demonstrate that the explicit shape encoding of GCNs leads to fewer segmentation outliers.
    \item \textbf{Speed}: Model run-time is an important consideration for echocardiography measurement tools since algorithms may be implemented on point of care systems with variable computational resources. We achieve decreased run-time relative to semantic segmentation using the GCN segmentation approach.
    \item \textbf{Landmark prediction:} The predicted graph nodes can be directly tied to important left ventricle landmarks such as the apex and basal points which are required for volume estimation and other downstream tasks. Semantic segmentation requires an extra module or rules-based analysis to identify these landmarks. 
    \item \textbf{Related tasks are predicted by a single model:} Segmentation and EF measurement predictions are processed together based on a single multi-frame encoder. Such an approach leads to improved performance, accuracy (since these tasks contribute to each other), and interpretability of clinical predictions.  
\end{itemize}

\section{Method}
\label{sec:Method}

Ejection fraction estimation requires the prediction of the left ventricle endocardial border at two points in the cardiac cycle: end-diastole (ED) and end-systole (ES). Sec.~\ref{subsec:BuildingBlocks} describes the building blocks of the proposed approach while Sec.~\ref{subsec:SegandEFPrediction} describes how those blocks can be combined for segmentation and EF prediction.

\subsection{Building blocks}
\label{subsec:BuildingBlocks}
Our method consists of four components: a CNN encoder that outputs a feature vector, a decoding graph that regresses the keypoints based on the feature representations, and two regression layers that directly outputs a value for the ejection fraction or a classification of the frame. The proposed method is completely modular and different networks can be substituted for each component.

\begin{figure}[t]
    \centering
    \includegraphics[width=\textwidth]{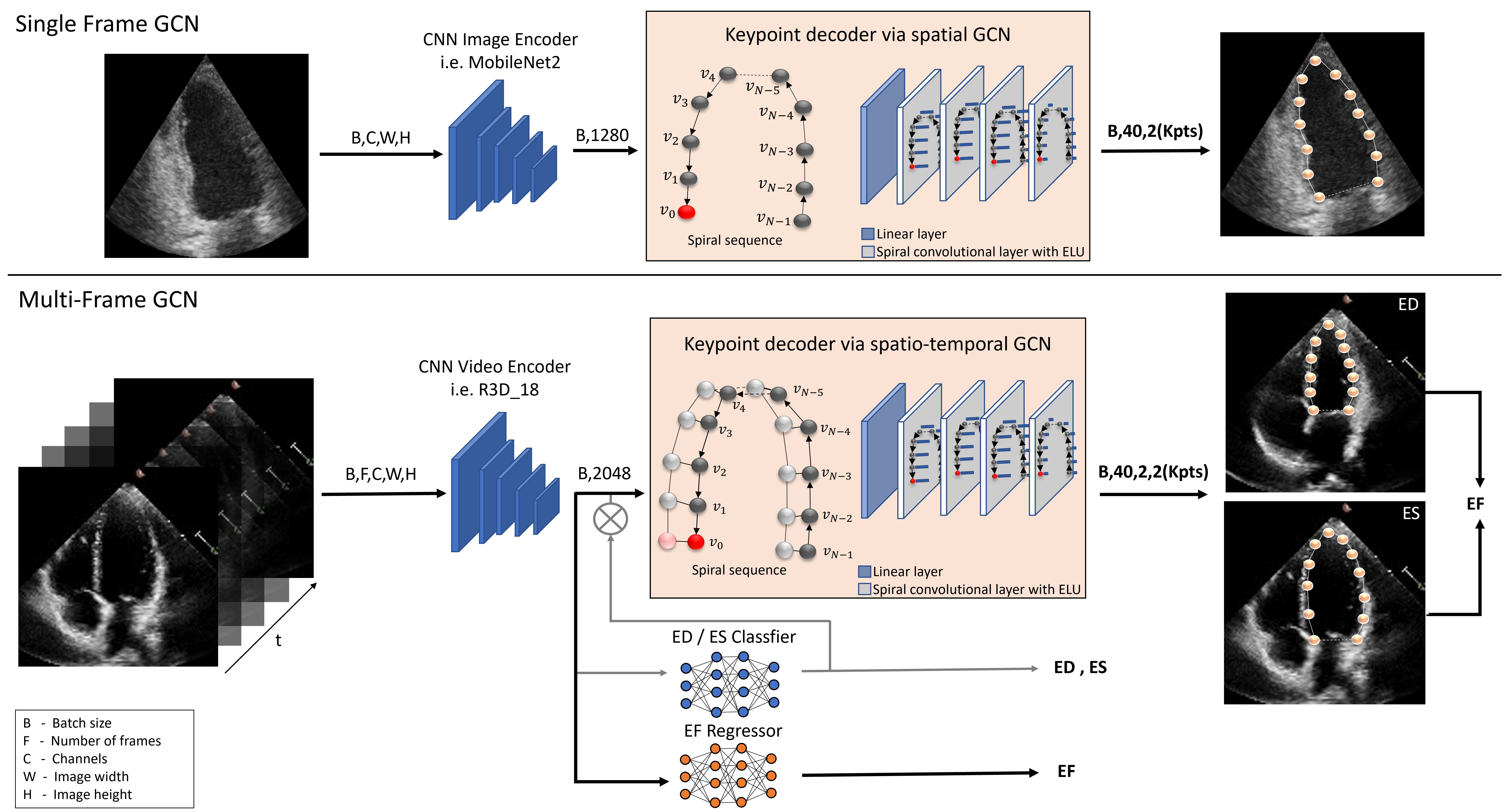}
    \caption{\textbf{Overview of the proposed method} Single frame GCN (top): The output of the CNN encoder is forwarded to the GCN with four spiral convolution layers that regresses a coordinate vector for each keypoint.
    Multi-frame GCN (bottom): The image sequence consisting of $F$ frames is fed into a CNN video encoder that outputs a feature representation. This feature representation is passed to a regressor module to predict the EF value and the spatio-temporal GCN that outputs keypoint coordinates for the two key frames. An ED/ES classifier is optionally added when the location of those frames is unknown (Sec.~\ref{subsec:SegandEFPrediction})} 
    \label{fig:overview}
\end{figure}

\subsubsection{Graph convolutional point decoder}
\label{subsec:GraphConvolutionalNetwork}
The left ventricle is an anatomical shape that can be described by a closed contour and approximated by a finite number of points sampled from the contour. Given a single ultrasound frame, we can interpret the contour of the left ventricle as an undirected graph $G = (V,E)$. Nodes $V =\{v_i | i =1, ...,N\}$ represent the contour points and $E$ represent the connections between the points. Spatial changes of one contour point most likely affect the neighboring points but may also affect other points that are not in direct proximity. 
Graph convolutional neural networks (GCNs) can be seen as a generalization of CNNs that do not require a fixed grid but can operate on any non-Euclidean structured data by aggregating the information of the neighbors using the edges and applying a weighting term \cite{gong2019spiralnet}. 
While several graph network models can be used, the proposed method relies on spiral convolutions \cite{bouritsas2019neural}. This operator enforces a fixed ordering of neighboring nodes during message passing to compute node updates with $x^k_i = \gamma^k  \left( \parallel_{j\in S(i,l)}  x_j^{k-1} \right)$ where $S(i,l)$ is the fixed spiral concatenation of the neighboring nodes $x_j$ and $\gamma$ is a multi-layer perceptron. Spiral connections were chosen because they can explicitly model an inductive bias of neighboring nodes and are computationally efficient. Although the approach was inspired by closed mesh structures and the spiral was defined on the mesh vertices, it can be also applied to circular structures like the left ventricle by defining the starting index and then generating the sequence in a clock-wise manner. 
The proposed graph decoder architecture consists of one initial dense layer that further compresses the input feature vector followed by four spiral convolution layers that are complemented by an exponential linear activation unit. In the last layer, two values are predicted for each node that represent the coordinate pixel position in the image.

\subsubsection{Encoder}
In the input layer of the graph, each node is assigned with a feature representation. From the original image or video information, a distinct lower dimensional feature vector must be extracted to meet these input requirements. For that purpose, any CNN network can be used and the choice of architecture is a trade-off between speed and accuracy. Each node is assigned with the CNN output concatenated with the neighboring node features following the order of the spiral sequence.

\subsubsection{EF regressor}
\label{subsec:EFRegressor}
The ejection fraction is defined as $EF = (EDV-ESV)/EDV$ where EDV and ESV are the volumes at end-diastole and end-systole respectively. While the EF can be estimated from the extracted keypoints using the Simpson biplane method, following earlier work \cite{kazemi2020deep,Ouyang,reynaud2021ultrasound}, we found that more accurate EF prediction could be obtained from a direct estimation. Our method therefore generates one EF prediction from keypoint predictions and one from the EF regressor. This approach yields accurate results from the direct regression along with a confidence check and explainable results from the keypoints. The proposed method uses a 4-layer multi-perceptron network (each linear layer is followed by an exponential linear activation unit), which takes the feature vector from the encoder and outputs a single predicted EF value. 

\subsubsection{ES/ED classifier}
EF estimation from an unlabeled video requires estimation of the ES and ED frames to know where segmentation should be applied. For predicting the ES and ED frame indices we use another multi-perceptron network based on 4 linear layers, in which each layer is followed by normalization and ReLU activation. Similar to~\cite{reynaud2021ultrasound}, the network outputs two arrays that represent the likelihood of each frame to be the ES or ED. A weighted cross-entropy loss is then applied to match each array against the ground truth index location.

\subsection{EchoGraphs - Left ventricle segmentation and EF prediction}
\label{subsec:SegandEFPrediction}

\subsubsection{Single frame segmentation}
An image encoder and graph point decoder can be used to predict the pixel location $(x,y) \in R^2$ of each keypoint $v_i \in V$ given a single image frame as shown in the top part of Fig. \ref{fig:overview}. Based on the shape constraint enforced by the GCN configuration, the output is a closed contour along the myocardial border that represents a segmentation of the left ventricle.

\subsubsection{Multi-frame segmentation and EF prediction with known ED/ES}
To measure EF, multiple input frames must be analyzed. We extend the proposed single frame approach discussed above to optimize keypoint contours on both ED and ES frames. One key property of the sequence is the shape consistency of the left ventricle between consecutive frames.
Taking a similar approach to previous work on body pose estimation \cite{yan2018}, each graph node at ED is connected to the corresponding node in ES to model the temporal connections. Adding these temporal edges between two consecutive frames can help to enforce consistency between the predicted segmentations. 

For this approach, the R(2+1)D video encoder \cite{tran2018closer} is used for generating the feature vector. This network takes as input a sequence of 16 frames where the first frame is ED and the last is ES. Including the intermediary frames gives better results than using a two frame approach with only ED and ES, indicating the extra context helps with learning. This approach requires manual identification of ED and ES frames. The output of the encoder serves as the input of the EF regression network and the graph decoder. EF prediction in this approach comes from both the keypoints and an EF regressor attached to the feature vector. 

\subsubsection{Multi-frame segmentation and EF prediction from unknown ED/ES}
If ED and ES are unknown, we demonstrate two possible approaches. First, the above approaches can be combined in a two-stage solution where the single-frame model first predicts keypoints for each frame. Those keypoints allow the computation of ventricle volumes and following the approach of \cite{Ouyang}, negative and positive peaks of the volumes can be identified. Subsequently, the multi-frame model can be applied to each peak pair to estimate the EF.

In the second approach, an ED/ES classifier is added that takes the encoder output and predicts the occurrence of the ED/ES frames in the sequence. The classifier output is concatenated with the feature vector from the encoder as input for the graph decoder. If a ED or ES frame is detected, the output of the decoder resembles the respective ventricle contours.
During inference, a sliding window approach is applied to process the entire video and EF results are averaged across the sequence. The multi-frame approaches with/without the ED/ES classifier are shown at the bottom of Fig. \ref{fig:overview}.

\section{Experiments \& Results}

\label{sec:ExperimentsResults}
Following research questions were asked in this project:
\begin{itemize}
    \item [Q1] How accurate, efficient, and robust is segmentation of a single frame using GCNs? (sec. \ref{subsec:Segmentation})
    \item [Q2] Given a single heart cycle video with labeled ED and ES frames, can we predict EF along with keypoints on the ED and ES frames? (sec. \ref{subsec:EFPrediction})
    \item [Q3] Given an unlabeled video with one or more cycles, can we predict the EF and ED/ES keypoints? (sec. \ref{subsec:EFPrediction})
\end{itemize}

Experiments to answer these questions are detailed below. Implementation details are in the supplementary material and source code to reproduce the given results is on Github: \url{https://github.com/guybenyosef/EchoGraphs} .

\subsection{Dataset}
\label{subsec:ValidationDatasets}
All Echographs models were trained and evaluated on the EchoNet open-access dataset \cite{Ouyang}. The dataset consists of 10,030 echocardiography videos of healthy and pathological patients. 
All videos have two frames labeled - the ED and ES frame for one selected cycle. The 40 annotated keypoints provide an approximation of the left ventricle contour plus 2 additional keypoints for the basal and apex point. Furthermore, the dataset provided EF values for each sequence. 
The dataset authors provide splits of the data into $80\%$ training, $10\%$ validation and $10\%$ test sets to allow direct comparison. Evaluations of all methods were performed on the same independent test set not seen during training.

\begin{figure}[t]
    \centering
    
    \includegraphics[width=\textwidth]{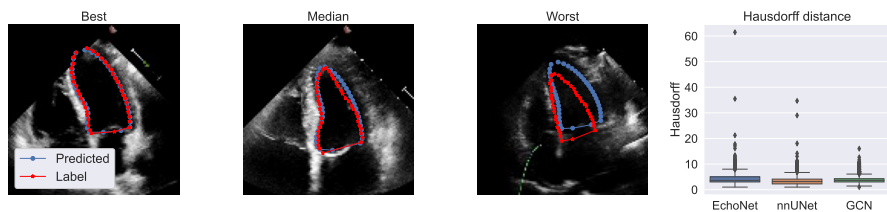}
    \caption{\textbf{Left:} Best, median, and worst results from the EchoGraphs single-frame segmentation network. The model prediction in the worst case appears to be more accurate than the label. \textbf{Right:} Hausdorff distance (in pixels) box plot of segmentation results. Compared to semantic segmentation, the EchoGraphs (with MobileNet2 backbone) is more robust.} 
    \label{fig:results}
\end{figure}

\subsection{Segmentation} 
\label{subsec:Segmentation}
To answer Q1 the single-frame model approach (Sec. \ref{subsec:SegandEFPrediction}) was trained on all annotated frames from the EchoNet dataset. Three different encoder backbones were tested for the EchoGraphs model to evaluate the trade-off between runtime and accuracy: MobilenetV2~\cite{sandler2018mobilenetv2}, ResNet18, and ResNet50~\cite{he2016deep}. We compare to two semantic segmentation algorithms: the DeepLabv3 approach from EchoNet~\cite{Ouyang}  and the nnU-Net~\cite{isensee2021nnu}. Results are shown in Table~\ref{tab:segmentation} where all approaches were compared on Dice score, mean keypoint error, performance, and memory size. The EchoNet dataset does not have real image sizes so to calculate keypoint error, coordinates were normalized by image dimensions. Example contours are shown in Fig. \ref{fig:results}. Dice results were all found to be significantly different using a Wilcoxon signed-rank test ($p_{wilcoxon}\ll 0.01$).

\begin{table}[]
\centering
\begin{tabular}{@{}llllllll@{}}
\toprule
Model                          & Backbone  & Dice (\%)              & MKE (\%) & Runtime [cpu/gpu] & Parameters \\ \midrule
EchoNet \cite{Ouyang}         & DeepLabV3 & $91.7\pm4.2$   &  $2.5\pm1.2$                  & 33.65/4.94               & 39.6 M     \\
nnU-Net \cite{isensee2021nnu} & U-Net     & $\mathbf{92.8\pm 3.6}$  &  $2.3\pm1.2$                  & 14.86/1.05           & 7.3 M      \\
EchoGraphs (ours)                     & MobileNetv2 &  91.6 $\pm$ 4.0  & $2.3\pm1.0$                 & {\bf 2.45}/0.68           & {\bf 4.92 M}     \\
EchoGraphs (ours)                     & ResNet18  & $91.8 \pm 4.0$ & $2.3\pm1.0$                    & 2.68/{\bf 0.46}           & 12.1 M     \\
EchoGraphs (ours)                     & ResNet50  & $92.1\pm 3.8$  & $\mathbf{2.2\pm0.9}$    &  6.73/1.05          & 27.1 M     \\ \bottomrule
\end{tabular}
\caption{Segmentation accuracy and performance for different methods evaluated on the EchoNet \cite{Ouyang} test set with 1264 patients and two annotated frames (ED and ES) each. MKE = mean keypoint error (mean L1 error in \%). Runtime is measured in msec per frame for a single forward pass of the model without preprocessing or augmentation.}
\label{tab:segmentation}
\end{table}

\subsection{EF prediction}
\label{subsec:EFPrediction}
The approaches described in Sec. \ref{subsec:SegandEFPrediction} were applied to the problem of EF prediction to answer Q2/Q3. Our methods were compared to the best results given by \cite{Ouyang} and \cite{reynaud2021ultrasound}. Results are shown in Table \ref{tab:ef-results} and all approaches are compared on mean absolute error (MAE), root mean squared error (RMSE), and correlation ($R^2$) averaged over all patients. 

\begin{table}[]
\centering
\begin{tabular}{@{}p{5.6cm}lclll@{}}
\toprule
Method                  & Input            & Frames & MAE & RMSE & $R^2$ \\ \midrule
EchoNet (MC3) \cite{Ouyang}             & single heartbeat    & 32         &  4.22  &  5.56    &    0.79   \\
Transformer (M.) \cite{reynaud2021ultrasound}             & single heartbeat             & 128         & 5.32    &  7.23    &  0.64     \\
Regression only (ours)   & single heartbeat & 16 & 4.28 &  5.75 & 0.72 \\
EchoGraphs (ours) using keypoints   & single heartbeat & 16         &  4.66   & 6.30     & 0.73    \\
EchoGraphs (ours) with regression head   & single heartbeat   & 16         &  \textbf{4.01}   & \textbf{5.36}      &  \textbf{0.81 }     \\ \midrule
EchoNet \cite{Ouyang}       & whole video      & -          &  \textbf{4.05}     & \textbf{5.32}     &  \textbf{0.81}   \\
Transformer \cite{reynaud2021ultrasound}            & whole video      & -          & 5.95     & 8.38      &  0.52     \\
EchoGraphs (ours) - Peak computation  & whole video  & - & 4.30      &5.86     & 0.65  \\ 
EchoGraphs (ours) - ED/ES classifier  & whole video  & - & 4.23      & 5.67     & 0.79  \\ 

\bottomrule
\end{tabular}
\caption{EF prediction results of the EchoGraphs models evaluated on 1264 patients. The top section of the table shows results from prediction on a single heartbeat while the bottom shows performance given a video with one or more cycles. The multi-frame EchoGraphs allows EF prediction based on the keypoints and the direct regression. Both results are listed for the single heartbeat while only regression results are listed for the whole video. The EchoGraphs approach with peak computation required one initial run of the single-frame approach. In four cases no peaks could be found and the method failed. Only the best approaches from \cite{Ouyang} and \cite{reynaud2021ultrasound} are listed.}
\label{tab:ef-results}
\end{table}

\section{Discussion \& Conclusion}
\label{sec:Discussion}
In this work, we have proposed a graph convolutional neural network for segmenting the myocardial border of the left ventricle and predicting ejection fraction for echocardiography videos. For the task of segmentation, our EchoGraphs method reaches comparable accuracy results in Dice score and mean keypoint error to state-of-the-art methods while achieving considerably better run-time performance. These results show graph convolutional networks are a more suitable method for segmentation when performance is critical. In addition to being faster, EchoGraphs are more robust, as demonstrated by the Hausdorff distance box plots in Fig. \ref{fig:results}. Most failure cases could be attributed to low image quality or false annotations (Fig. \ref{fig:results} - worst). The robustness of the EchoGraphs can be attributed to the explicit shape encoding in the graph structure while the runtime is reduced because the decoding path is simpler.

Although segmentation accuracy is slightly below nnU-Net, the nnU-Net pipeline contains test-time augmentations and an abundance of hyperparameter optimizations based on dataset characteristics. These same optimizations could be applied to EchoGraphs and likely further improve performance. 

We achieve state-of-the-art results for EF prediction from echocardiography clips. We demonstrate an interpretable approach based on the volume estimation from the predicted keypoint segmentations, and a slightly more accurate black-box approach based on direct EF estimation from the encoder feature vector. Our results show that a GCN-based multi-task approach for simultaneously learning both keypoints and clinical measurements can achieve high accuracy for all tasks and outperforms direct regression ($p_{wilcoxon}\ll 0.01$). We present two approaches for EF prediction from videos and both approaches achieve results comparable to previous methods while simultaneously predicting left ventricle segmentations. The EF errors for each of our EchoGraphs models were significantly below the 7-13\% inter-observer error range reported in \cite{Ouyang}.

The proposed approach may provide more efficient and accurate measurement systems for commercial scanner systems. In the future, we aim to extend the approach to other clinical measures. The predicted keypoints provide important information that can directly be utilized for other applications such as foreshortening detection and strain analysis.

%

\bibliographystyle{splncs04}
\bibliography{paper2336}

\end{document}